\pdfoutput=1

\documentclass[11pt]{article}

\usepackage[final]{acl}

\usepackage{times}
\usepackage{latexsym}

\usepackage[T1]{fontenc}

\usepackage[utf8]{inputenc}

\usepackage{microtype}

\usepackage{inconsolata}

\usepackage{graphicx}
\usepackage{booktabs}
\usepackage[textsize=tiny]{todonotes}
\usepackage{amsmath}
\usepackage{amssymb}
\usepackage{mathtools}
\usepackage{amsthm}
\usepackage{multicol}
\usepackage{colortbl}
\usepackage{color}
\usepackage{pifont}

\usepackage[capitalize,noabbrev]{cleveref}

\theoremstyle{plain}

\theoremstyle{definition}

\theoremstyle{remark}

\definecolor{Red}{RGB}{176,36,24}
\definecolor{codeblue}{rgb}{0.25,0.5,0.5}
\definecolor{myblue}{rgb}{0.88,0.98,1}
\definecolor{mygreen}{rgb}{0.92, 1.0, 0.92}
\definecolor{myred}{rgb}{1, 0.9, 0.9}
\definecolor{mygray}{gray}{0.9}

\definecolor{Highlight}{HTML}{E8F8F5}
\definecolor{midgreen}{HTML}{69c5a3}
\definecolor{midblue}{HTML}{69a3f1}
\definecolor{darkgreen}{HTML}{146038}
\definecolor{darkblue}{HTML}{143b59}
\definecolor{red1}{HTML}{C00000}

\definecolor{LightCyan}{rgb}{0.88,1,1}

\definecolor{hotpink}{RGB}{59, 115, 227}

\newcommand{\x}{{\bf x}}

\newcommand{\y}{{\bf y}}

\newcommand{\eg}{\emph{e.g.}}
\newcommand{\ie}{\emph{i.e.}}
\newcommand{\name}{{\sc TVC }}
\newcommand{\mame}{{\sc TVC}}

\title{Mitigating Visual Forgetting via Take-along Visual Conditioning \\for Multi-modal Long CoT Reasoning}

\author{
  \textbf{Hai-Long Sun\textsuperscript{1,2}} \thanks{Work done during their internship/employment at Tencent Hunyuan},
  \textbf{Zhun Sun\textsuperscript{4,5}} \footnotemark[1],
  \textbf{Houwen Peng\textsuperscript{3}},
  \textbf{Han-Jia Ye\textsuperscript{1,2}} \thanks{Corresponding author: yehj@lamda.nju.edu.cn}
  \\
  \textsuperscript{1}School of Artificial Intelligence, Nanjing University\\
  \textsuperscript{2}National Key Laboratory for Novel Software Technology, Nanjing University\\
  \textsuperscript{3}Tencent 
  \textsuperscript{4}Center for Language AI Research, Tohoku University \\
  \textsuperscript{5}RIKEN Center for Advanced Intelligence Project \\
}

\begin{document}
\maketitle

\begin{abstract}
    Recent advancements in Large Language Models (LLMs) have demonstrated enhanced reasoning capabilities, evolving from Chain-of-Thought (CoT) prompting to advanced, product-oriented solutions like OpenAI o1. 
    During our re-implementation of this model, we noticed that in multimodal tasks requiring visual input (\eg, geometry problems), Multimodal LLMs (MLLMs) struggle to maintain focus on the visual information, in other words, MLLMs suffer from a gradual decline in attention to visual information as reasoning progresses, causing text-over-relied outputs.
    To investigate this, we ablate image inputs during long-chain reasoning. Concretely, we truncate the reasoning process midway, then re-complete the reasoning process with the input image removed. We observe only a $\sim$2\% accuracy drop on MathVista’s test-hard subset, revealing the model's textual outputs dominate the following reasoning process. 
    Motivated by this, we propose Take-along Visual Conditioning (TVC), a strategy that shifts image input to critical reasoning stages and compresses redundant visual tokens via dynamic pruning. This methodology helps the model retain attention to the visual components throughout the reasoning. Our approach achieves state-of-the-art performance on average across five mathematical reasoning benchmarks (+3.4 points vs previous sota), demonstrating the effectiveness of TVC in enhancing multimodal reasoning systems.
    The project page is available at \url{https://sun-hailong.github.io/projects/TVC}. \looseness=-1
\end{abstract}
\section{Introduction}
 
Large Language Models (LLMs) have achieved significant advancements in natural language processing (NLP), particularly in the area of reasoning. 
These models have evolved from simple prompt-based Chain-of-Thought (CoT)~\citep{wei2022chain} techniques to sophisticated product-oriented solutions like OpenAI's o1~\citep{openai2024o1}, DeepSeek-R1~\citep{deepseek2024r1}, and Qwen-QVQ~\citep{alibaba2024qvq}, demonstrating iterative reasoning capabilities for complex multi-step tasks, which enables them to handle tasks that require multi-step thinking, logic, and knowledge integration. Recently, several works also extended the CoT reasoning process to MLLMs settings through data-centric innovations. For instance, Math-LLaVA~\citep{shi2024math} pioneers domain-specific training with the MathV360K dataset, while MAmmoTH-VL~\citep{guo2024mammothvl} scales up multimodal CoT data generation. \looseness=-2

While such progress is notable in text-based domains, extending these advancements to multimodal scenarios presents unique challenges that transcend traditional language model boundaries. Reasoning in MLLMs requires fused understanding across different modalities, for example, in geometric reasoning tasks the model should interpret and reason about images alongside text. Therefore, the model’s ability to integrate and maintain focus on both types of information is critical.
Unlike text-only LLMs that reinforce problem context through linguistic recurrence, MLLMs struggle to sustain visual attention across reasoning steps. That is, as the length of the context increases, the model is more inclined to conduct the next step of reasoning based on the previously outputted text rather than the information of the image itself, which eventually leads to the continuation of the wrong text reasoning process and degraded model performance. We term this phenomenon as visual forgetting.

In this work, we conduct a diagnostic analysis of the visual forgetting effect within a long-chain reasoning system. The system processes multimodal Q\&A tasks through a series of interconnected reasoning steps. We demonstrate significantly reduced attentional allocation to visual inputs during multi-stage reasoning processes. More formally, our analysis: 1) truncates the reasoning process midway and removes the image embeddings; 2) regenerates subsequent reasoning trajectories; 3) evaluates the reasoning outcomes of the pre/post-ablation inference trajectories. Intuitively, the performance gap between normal reasoning and diagnostic reasoning reveals the model's dependency on generated text over original visual evidence. Our results (See \Cref{sec:visual_forgetting}) on the MathVista-Hard datasets show that, removing the image midway through the reasoning only causes an insignificant performance degradation (\ie $\sim$2\%), indicating that the model completes the reasoning process based primarily on its output text. More importantly, we also observe that the model’s dependency on the visual evidence diminishes over time since the early removal of the image inputs could hurt accuracy by $\sim$20\%. This suggests model's reasoning employs both visual and textual information in the early stage, then over-relying on text history which limits full visual reasoning potential.

Motivated by this, we propose a novel strategy to mitigate the visual forgetting effect and maintain visual attention throughout the reasoning process. Our methodology compresses and shifts the image input to the later stages of the reasoning process, ensuring the model integrates sufficient visual evidence into its reasoning. This approach results in improved performance and achieves state-of-the-art results on average across five mathematical reasoning benchmarks (\ie, +3.4\% vs previous sota). Our findings highlight the effectiveness of this strategy in enhancing the performance of multimodal reasoning systems, providing a robust solution to the problem of visual forgetting in long-chain reasoning tasks. \looseness=-1

\vspace{-1pt}
\section{Take-along Visual Conditioning: Sustaining Visual Evidence for Multi-modal Long CoT Reasoning}
\vspace{-1pt}
In this section, we first discuss our motivation and observations of the visual forgetting phenomenon in MLLM reasoning systems (Section~\ref{sec:visual_forgetting}). Then, we propose the Take-alone Visual Conditioning (\mame) approach to mitigate visual forgetting and enhance the model's long-chain reasoning capabilities (Section~\ref{sec:TVC}).

\begin{figure}[t]
\centering
\includegraphics[width=.9\linewidth]{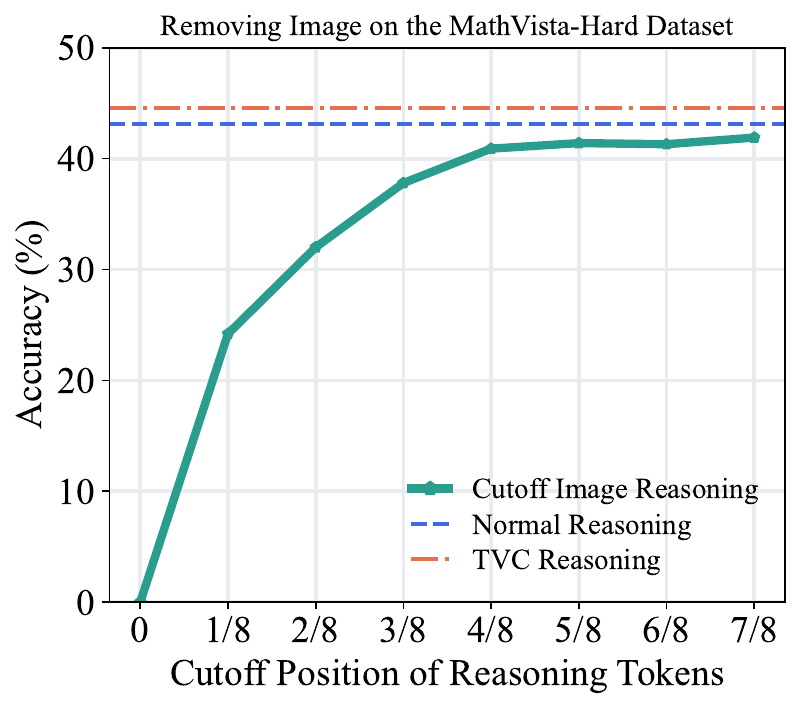}
\caption{\textbf{The visual forgetting phenomenon by removing the image at different reasoning stages.} It shows that by the midpoint of the reasoning process, the model becomes less dependent on the image, causing text-over-relied outputs.}
\label{fig:visual_forgetting}
\end{figure}

\subsection{Capturing the Visual Forgetting}
\label{sec:visual_forgetting}

Text-based reasoning systems often repeat key terms or phrases (\eg, restating ``triangle ABC'' or ``equation (1)'') to keep the problem's context clear and focused. By restating these critical details, they create strong connections between each step of the reasoning process and the original problem’s requirements. This repetition ensures the logic stays on track, and consistent to follow.
\begin{equation}
\label{eq:mllm}
\mathcal{C}_{\text{MLLM}} = f(V, T_1, ..., T_n) 
\end{equation}
On the other hand, MLLMs struggle with this approach due to their design. As formalized in \Cref{eq:mllm}, visual inputs $V$ are confined to the initial processing stages. Unlike text, these visual evidence aren’t revisited or reinforced later in the reasoning process. Because there’s no built-in way to keep visual information ``active'' or relevant throughout the task, the system’s ability to connect visual details with text or logic weakens over time, leading to a progressive \textit{visual attention decay}. The model is more likely to reason with previously outputted text and becomes particularly problematic in visual reasoning tasks that require continuous validation of spatial relationships.

We conduct two analytic analyses using the QVQ-72B-Preview model~\citep{alibaba2024qvq} to capture this visual forgetting phenomenon quantitatively and qualitatively. On one hand, we remove the visual inputs at eight different stages to observe their impact. On the other hand, we depict the attention matrix to directly track the attention decay of the visual evidence over time.

\noindent\textbf{Progressive Image Removing.}
To assess the extent to which the reasoning process depends on previously generated text, we first perform a normal reasoning process, then reset the KV cache at various stages of the reasoning process. This effectively removed image tokens and forced subsequent steps to rely solely on text-based information. 
Specifically, the reasoning process was divided into $K=8$ intervals based on output token counts regardless of the length of the normal reasoning process, with visual input progressively masked by resetting the KV cache at different cutoff positions $k \in \{0, 1, \dots, K-1\}$. In other words, the first $k/8$ part of the normal reasoning process is now employed as a prompt, and the model now re-complete the reasoning process without image inputs.
Furthermore, we discovered that for some questions (30.9\% of the MathVista dataset), the model could answer correctly using only the text-based prompt. Consequently, we excluded these questions and designated the remaining dataset as the \emph{MathVista-Hard} dataset.

\begin{figure}[t]
\centering
\includegraphics[width=.9\linewidth]{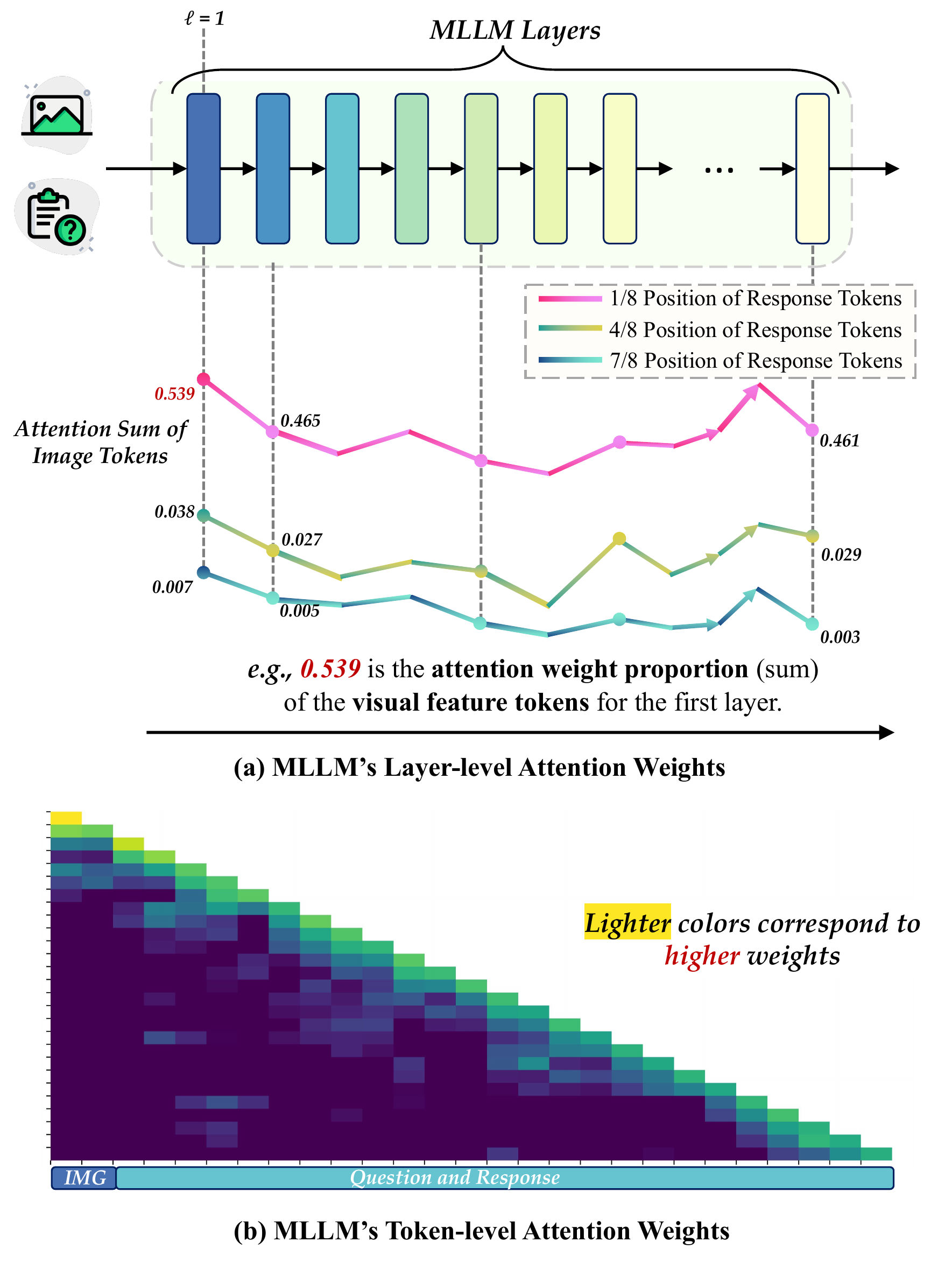}
\caption{\textbf{Illustration of layer-level and token-level attention weights.} (a) The layer-level attention weights of image tokens across different response token positions. (b) The token-level attention weights at the middle layer. It shows that the model's attention to the image gradually decreases during the reasoning process. }
\label{fig:visual_attention}
\end{figure}

As illustrated in Figure~\ref{fig:visual_forgetting}, we observe that the performance is 40.9 at $k = 4$ and 43.1 for the normal reasoning. This minimal 2.2\% gap suggests that the model overly relies on text outputs during the later reasoning stages rather than on the image.
This indicates that once the model completes the half stages of reasoning, it becomes overly reliant on its own generated CoT. Over-reliance on this self-generated reasoning worsens the model's tendency to forget visual evidence over time, which can result in hallucinations during lengthy, multi-step reasoning tasks.
This result also reveals a near-exponential forgetting effect in performance as $k$ increases, which we formalize as:  \looseness=-1
\begin{equation}
    \mathcal{R}(k) = \mathcal{R}_{\text{full}} - \Delta_{\text{visual}}(k), \quad \Delta_{\text{visual}}(k) \propto e^{-k},
\end{equation}
where $\mathcal{R}_{\text{full}}$ represents the full multimodal reasoning performance, and $\Delta_{\text{visual}}(k)$ quantifies the performance degradation caused by visual masking at position $k$.

\noindent\textbf{Visual Attention Decay.}
To more intuitively observe the changes in visual feature attention across different stages of reasoning, we investigate the layer-level attention weights at various response token positions and the token-level attention weights at layer 16. Given that MLLM’s attention weights reflect the focus on tokens and influence the decision-making process, we first analyze the attention weights at each layer of the MLLM. Specifically, for each layer, we calculate the proportion of attention weights on all image tokens. \looseness=-1

As shown in Figure~\ref{fig:visual_attention}(a), we observe that at the 1/8 position of the reasoning process, the model effectively focuses on the visual inputs. However, as reasoning progresses, despite fluctuations in attention to visual tokens across layers, the model's overall attention to visual evidence gradually decreases, leading to visual forgetting. Next, following the methodology of FastV~\cite{chen2024image}, we analyze the attention maps for several cases and find that the model predominantly focuses on previously generated text tokens rather than the input image. After approximately 20\% tokens, the existence of image inputs on attention maps diminishes significantly, as illustrated in Figure~\ref{fig:visual_attention}(b). Both of the observations indicate a phenomenon of visual memory degradation, revealing the model's limitations in maintaining consistent attention to visual inputs throughout the reasoning process.

\subsection{Take-along Visual Conditioning}
\label{sec:TVC}

In this section, we introduce our solution to tackle this problem in detail. We propose Take-along Visual Conditioning (\mame), a dynamic image retention mechanism that re-introduces visual inputs at strategic intervals throughout the reasoning process. \name mitigates visual attention decay by periodically reaffirming visual information, akin to human problem-solving behaviors where individuals frequently refer back to visual inputs. Our approach enhances the model's ability to incorporate visual information continuously, improving its long-chain reasoning capacity by ensuring that visual evidence is revisited during critical decision-making moments.

\begin{figure}[t]
\centering
\includegraphics[width=1.0\linewidth]{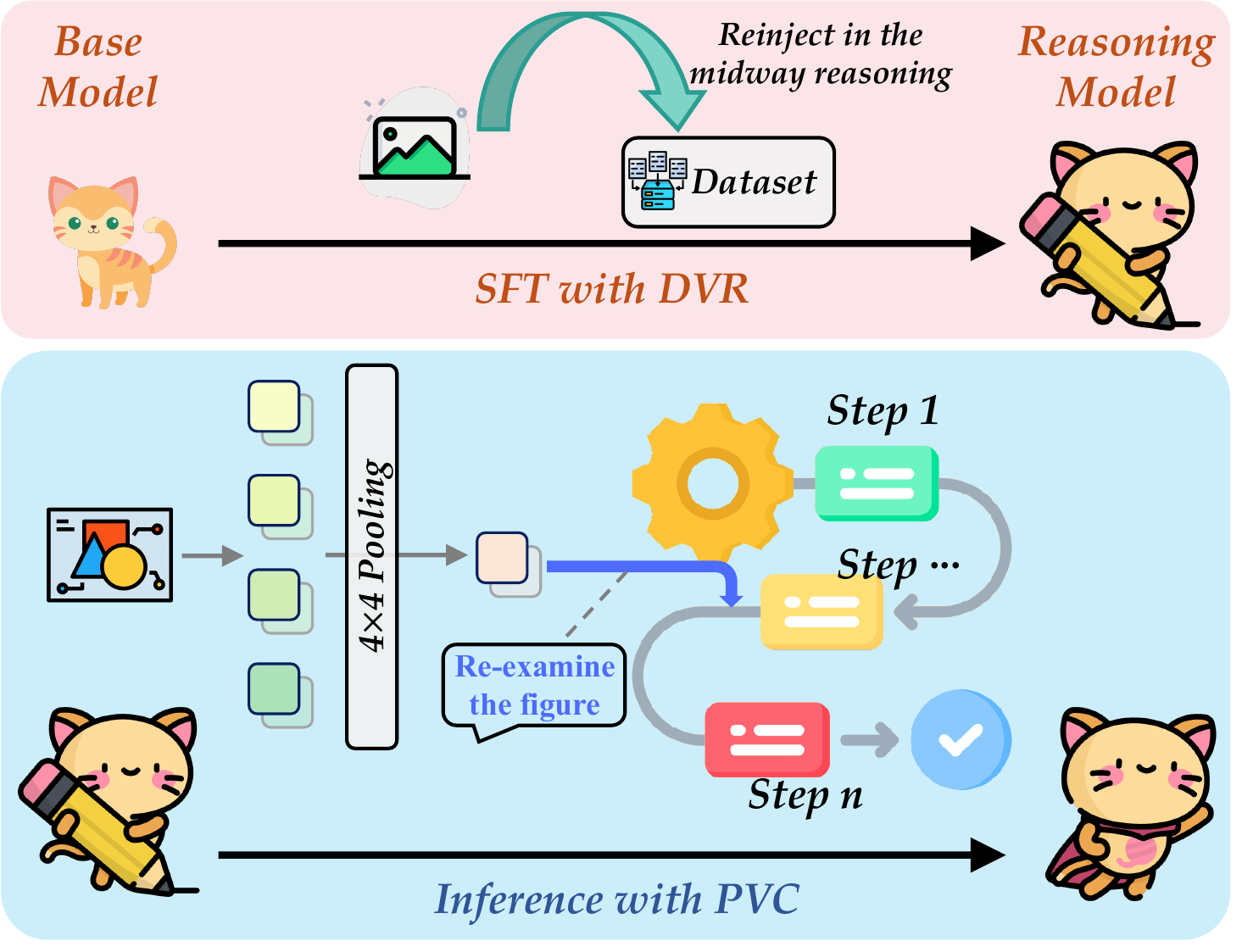}
\caption{\textbf{Overview of TVC System Design.} We enable the model to have take-along visual conditioning capabilities through two stages: training and inference.}
\label{fig:teaser}
\end{figure}

The \name method consists of two key stages: training and testing. In the training stage, we introduce Dynamic Visual Reaffirmation (DVR), which guides the model through iterative reinforcement of visual evidence during long reasoning chains. In the testing phase, we present Periodic Visual Calibration (PVC), where visual reactivation is periodically triggered at self-reflection intervals. To prevent the model from forgetting previous text-based reasoning steps due to an excessive number of image tokens, we adopt image compression through adaptive pooling to reduce the image token size while preserving spatial semantics. This dual-modality engagement mechanism ensures consistent interaction between textual reasoning and visual evidence. 
We present an illustration of the \name system in Figure~\ref{fig:teaser}.

\noindent\textbf{Dynamic Visual Reaffirmation.}
Our dynamic visual reaffirmation training strategy combines two key components:  
1) \textit{Data Curation}: We curate long-chain reasoning data using the pipeline described in Section~\ref{sec:data_pipeline}, sourced from high-quality academic datasets (\eg, MathV360K, Geo170K, and LLaVA-OneVision). This process yields a high-quality dataset optimized for long-chain reasoning training. 
2) \textit{Visual Content Injection}: While the curated data ensures correctness, the QVQ model inherently lacks the ability to iteratively reference visual inputs during reasoning. Therefore, we manually re-inject the visual content (visual embeddings and bridging prompt) to triggers visual re-activation at predefined self-reflection intervals. Specifically, given the initial multimodal input $\mathcal{M}_0 = (V, T_0)$, DVR performs visual reactivation at self-reflection intervals $\{r_1, ..., r_m\}$:  
\begin{equation}  
\mathcal{M}_i = \left(V,\ [T_\text{prev}; \underbrace{\text{Prompt}}_\text{Re-activation}; T_\text{new}]\right) \\ \text{at step } r_i  
\end{equation}  
where $T_\text{prev}$ represents the previous reasoning steps and $T_\text{new}$ denotes the new reasoning steps that are based on prior reasoning and reintroduce focus on visual information. The bridging prompt is employed to hint the existence of the image, \eg, Let me see the image again. To improve efficiency, our initial implementation adopts midpoint reactivation ($m=1, r_1=0.5L$ for $L$-step chains).

During the self-reflection phase of the reasoning steps, we randomly select $m$ positions to inject visual content. Specifically, we inject reactivation prompts and regenerate visual captions using the model’s intrinsic capabilities. This forces joint attention to both textual reasoning and visual evidence. By leveraging the model’s intrinsic image captioning capability, we continuously reinforce visual information throughout the reasoning process. This ensures that the model incorporates image evidence during reflection, rather than relying solely on textual reasoning.

\noindent\textbf{Periodic Visual Calibration.}  
Calibrating visual attention plays a crucial role in enhancing long-chain reasoning capabilities. After training our model, we further design the periodic visual calibration process. Specifically, we coordinate operations during reactivation as follows:
1)  \textit{Token Compression}: We first compress visual tokens using average pooling to prevent text-based reasoning from forgetting visual information. 
2) \textit{Visual Cache Reset}: We then prepend an instruction (bridging prompt) to re-introduce the image and re-inject image tokens by resetting the KV cache of the generation process.

We also provide an example of how PVC is implemented in the case study section (\Cref{sec:case_study}). PVC both improves reasoning efficiency and prevents the model from forgetting previous reasoning steps due to an overload of visual tokens.

\section{Data-Centric Implementation of Multimodal Reasoning System}
\label{sec:data_pipeline}
In this section, we briefly describe our implementation of the multimodal reasoning system through a scalable curated data generation pipeline.

\subsection{Long-Chain Reasoning Data Collection}
Prior research~\cite{qin2024o1,jiang2024technical} has identified two dominant paradigms for constructing long-chain reasoning systems: (1) explicit search-based methods, which utilize structures such as Monte Carlo Tree Search (MCTS) combined with specially trained reward models to guide the search process toward optimal solutions, and (2) instruction distillation approaches, which fine-tune models on curated long chain-of-thought (CoT) datasets. To efficiently develop an MLLM with long-chain reasoning capabilities, we adopt the distillation paradigm. In this section, we describe the distillation process and present the data generation pipeline aimed at enhancing the reasoning capability of MLLM.

\begin{figure}[t]
\centering
\includegraphics[width=0.98\linewidth]{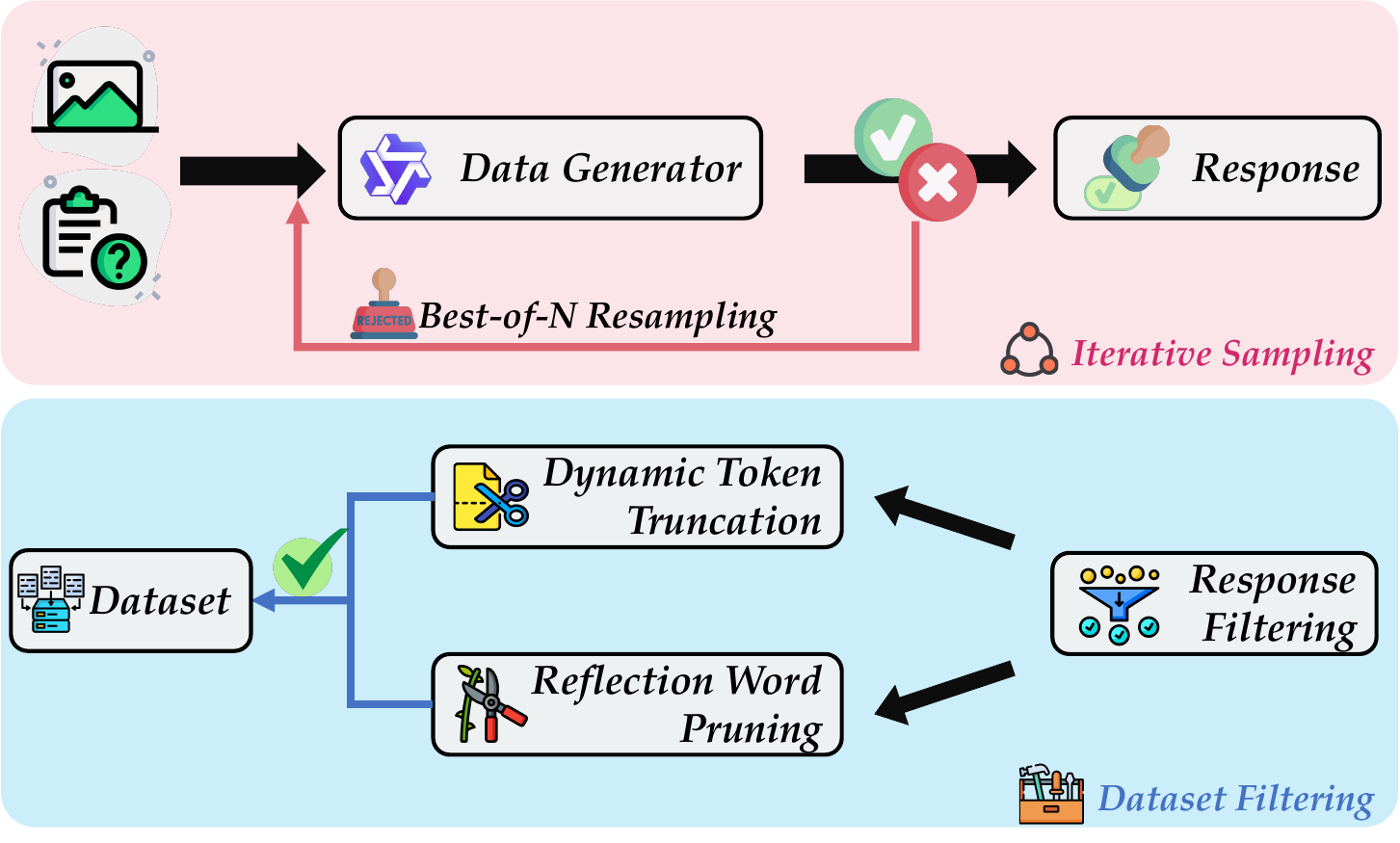}
\caption{\textbf{Data Generation Pipeline of TVC.} We use iterative distillation to collect long-chain reasoning data, followed by a comprehensive response filtering process to ensure high-quality reasoning.}
\label{fig:data_pipeline}
\end{figure}

Our distillation pipeline begins with aggregating publicly available reasoning datasets (\eg, MathV360K~\cite{shi2024math} and Geo170K~\cite{gao2023g}). Through an iterative process of distillation and response filtering, we progressively enhance the model's long-chain reasoning capabilities. Specifically, the teacher model, QVQ-72B-Preview, generates long COT reasoning data, while the student model, Qwen2-VL, undergoes supervised fine-tuning on the filtered CoT data to enhance its reasoning performance.

\subsection{Iterative Distillation with Reject Sampling}
After obtaining the long-chain reasoning responses, we employ an assessment pipeline to ensure data quality. Specifically, we introduce a dual-temperature sampling mechanism to optimize data quality through variance exploitation.

\noindent{\bf Deterministic Initial Sampling.} For the first-stage data generation, we employ temperature $\tau=0$ to obtain the QVQ's most confident reasoning paths:  
\begin{equation}
    \mathcal{D}_{\text{init}} = \{ (\x, \y_{\text{CoT}}) \sim \mathcal{P}_{\text{QVQ}}(\cdot|\x; \tau=0) \},
\end{equation} 
where $\y_\text{CoT}$ represents the response generated by QVQ. This ensures that the model generates the most probable reasoning path for each input. Subsequently, we use these high-confidence responses as a foundation for further refinement in later stages of data generation.

\noindent{\bf Answer-Centric Reject Sampling.} 
To ensure the quality of the data, we employ an LLM-as-a-Judge approach for answer-centric reject sampling. Specifically, we use a strong LLM (\eg, Qwen2.5-72B-Instruct) as the verifier and prompt the model to output a `yes' or `no' in JSON format, indicating whether the long-chain reasoning answer is correct, \ie, $\mathcal{Y}_{\text{valid}} = \{ \y_\text{CoT} | \text{Match}(\y_\text{CoT}, \y_{\text{gt}}) \}$. After this process, we obtain $\sim$200K high-quality samples.

\noindent{\bf Best-of-N Error Correction.}
While initial sampling with temperature $\tau=0$ yields a set of data, there are substantial residual errors ($\mathcal{D}_{\text{error}}$) where QVQ's responses ($\y_{\text{CoT}}$) unmatched with ground truth answers (55.8\% of $\mathcal{D}_{\text{init}}$). To fully leverage the available open-source data, we implement a contrastive regeneration strategy using $\tau=1$: 
\begin{equation}
\mathcal{D}_{\text{corrected}} = \bigcup_{\x \in \mathcal{D}_{\text{error}}} \left\{ \arg\min_{\y^{i} \in \mathcal{Y}_{\text{valid}}} \text{Length}(\y^{i}) \right\}_{i=1}^8,
\end{equation}
where $\mathcal{Y}_{\text{valid}}$ denotes the subset of 8 regenerated responses (at $\tau=1$) that satisfy answer correctness. For cases with multiple valid responses, we prioritize minimal reasoning token length to enhance conciseness and efficiency:
\begin{equation}
\text{Length}(\y) = \sum_{t=1}^T \mathbb{I}(w_t \notin \{\text{[PAD], [SEP]}\})
\end{equation} 
This two-stage sampling achieves partial error recovery while reducing the average token count to minimize meaningless self-reflection, effectively balancing correctness and efficiency.To further enhance the quality of the dataset, we also perform dynamic token truncation and reflection word pruning. This approach helps reduce the ineffective reflection parts in the dataset, thereby mitigating the issue of the model being unable to complete its answers. More details are in Section~\ref{sec:prune_dataset}.
\section{Experiments}

\begin{table*}[t]
  \centering
  \caption{\textbf{Results on Visual Reasoning Tasks.} We conduct evaluation experiments across 6 benchmarks, covering both general reasoning and task-specific reasoning assessments. \name exhibits notable effectiveness and generalizability when applied to Qwen2-VL, surpassing other state-of-the-art MLLMs by a large margin.}
    \resizebox{1.0\linewidth}{!}{
        \begin{tabular}{lcccccccc}
        \toprule
        Model & Size & MathVista & MathVision & MathVerse & Dynamath & OlympiadBench & Average \\
        \midrule 
        MiniCPM-V-2.6~\cite{yadav2025exploring} & 8B & 60.8 & 18.4 & 17.6 & 9.8 & -  & -\\
        VITA-1.5~\cite{fu2025vita} & 8B & 66.2 & 19.5 & 23.4 & 9.6 & - & -\\
        LLaVA-COT~\cite{xu2024llava} & 11B & 52.5 & 19.9 & 22.6 & 7.8 & - & -\\
        Qwen2-VL~\cite{Qwen2VL} & 7B & 60.9 & 16.3 & 24.6 & 11.0 & 3.2 & 23.2 \\
        InternVL2.5~\cite{chen2024expanding} & 8B & 64.5 & 17.0 & 22.8 & 9.4 & 0.1 & 22.8\\
        POINTS1.5~\cite{liu2024points1} & 8B &66.4 & 22.0 & 26.6 & 14.2 & - & -\\
        Ovis1.6-Gemma2~\cite{lu2024ovis} & 27B & 70.2 & 20.6 & 37.8 & 17.0 & -  & -\\
        InternVL2.5-COT~\cite{chen2024expanding} & 78B & 71.4 & 32.5 & 40.1 & 28.5  & -  & -\\
        LLaVA-OneVision~\cite{li2024llava} & 72B & 67.1 & 25.3 & 27.2 & 15.6 & -  & -\\
        Qwen2-VL~\cite{Qwen2VL} & 72B & 69.7 & 26.6 & 36.2 & 20.0 & 10.3 & 32.6 \\
        QVQ-72B-preview~\cite{alibaba2024qvq} & 72B & 71.4 & 35.9 & 41.5 & {\bf 30.7} & 20.4 & 40.0 \\
        \midrule
        \rowcolor{mygray} \mame & 7B & 68.1 & 22.7 & 38.9 & 15.1 & 9.8 & 30.9\\
        \rowcolor{mygray} \mame & 72B & {\bf 72.2} & {\bf 41.9} & {\bf 48.8} & 30.0 & {\bf 24.3} & {\bf 43.4} \\
        
        \bottomrule
        \end{tabular}%
    }
  \label{tab:main-exp} 
\end{table*}%

We conduct comprehensive experiments across multiple vision-language benchmarks to demonstrate the effectiveness of our method. Section~\ref{sec:imp_details} provides implementation details for \mame. In Section~\ref{sec:main_result}, we present key results on visual reasoning tasks, supplemented with findings on general image understanding. Section~\ref{sec:further_analysis} details ablation studies to evaluate the influence of critical design decisions. Finally, Section~\ref{sec:case_study} presents visualizations and case studies to illustrate the method's operational characteristics and insights derived from it.

\subsection{Training Recipe}
\label{sec:imp_details}
We integrate the \name system with MLLMs of varying scales to demonstrate the effectiveness and generalizability of our approach. Initial implementation with Qwen2-VL-7B-Instruct confirmed the method’s validity. To further evaluate scalability and establish robust baselines against state-of-the-art MLLMs, we expanded the approach to a 72B model.
Prior to training \mame, we follow the long-chain reasoning pipeline described earlier. We use the LLaMA-Factory~\cite{zheng2024llamafactory} framework, with a learning rate of 2e-5, a batch size of 256, and 5 training epochs. During optimization, only the LLM parameters and cross-modal connector are trained, while the visual encoder remains frozen. The training process requires 10 hours on a 64×H20 GPU setup for the 7B model and approximately 4 days for the 72B model. Additional details are presented in Table~\ref{tab:train_detail}.

\begin{table}[t]
    \centering
    \caption{\textbf{Ablations on the TVC System.} \name enhances reasoning capabilities, showing significant improvements on both general and task-specific reasoning benchmarks.}
    \resizebox{1.0\linewidth}{!}{
	\begin{tabular}{l|cccc}
        \toprule
		 Method&  MathVista & MathVision &  MathVerse & Avg  \\ 
         \midrule
		 Baseline&  60.9 & 16.3 & 24.6 & 33.9  \\
		 Vanilla - Direct SFT&  63.5 & 19.8 & 31.6 & 38.3  \\
		 TVC w/o PVC &  66.7 &  21.8 &  35.6 & 41.4  \\
		 TVC w/o DVR &  66.2 &  22.3 &  34.7 & 41.0  \\
		 \rowcolor{mygray} TVC Full&  68.1 & 22.7 & 38.9 & 43.2 \\
         \bottomrule
	\end{tabular}
    }
	
    \label{tab:ablation_component}
\end{table}

\subsection{Evaluation Setup} 
We conduct a comprehensive experimental analysis across various visual reasoning benchmarks that require advanced visual reasoning skills. To ensure a well-rounded evaluation, we select several widely recognized and representative benchmarks, including MathVista~\cite{mathvista}, MathVerse~\cite{zhang2024mathverse}, MathVision~\cite{wang2024measuring}, Dynamath~\cite{zou2024dynamath}, and OlympiadBench~\cite{he2024olympiadbench}. MathVista consists of 6,141 examples that require fine-grained, deep visual understanding and compositional reasoning. MathVerse contains 2,612 multi-subject math problems from a variety of sources. MathVision includes 3,040 high-quality mathematical problems sourced from established mathematics competitions. OlympiadBench features 8,476 bilingual multimodal problems tailored to Olympic-level mathematics and physics competitions. These benchmarks evaluate problem-solving abilities in mathematics, and following standard practice, we use GPT-4o-mini as the evaluator. Following the VLMEvalKit guidelines, we exclude the text-only split from MathVerse and the theorem-proof sections from OlympiadBench. For a fair comparison, we conduct evaluations using the testmini sets of MathVerse and MathVista. 
Fast evaluation is made possible through the use of the VLMEvalKit~\cite{duan2024vlmevalkit} and vLLM~\cite{kwon2023efficient}.

\begin{figure}[t]
\centering
\includegraphics[width=1.0\linewidth]{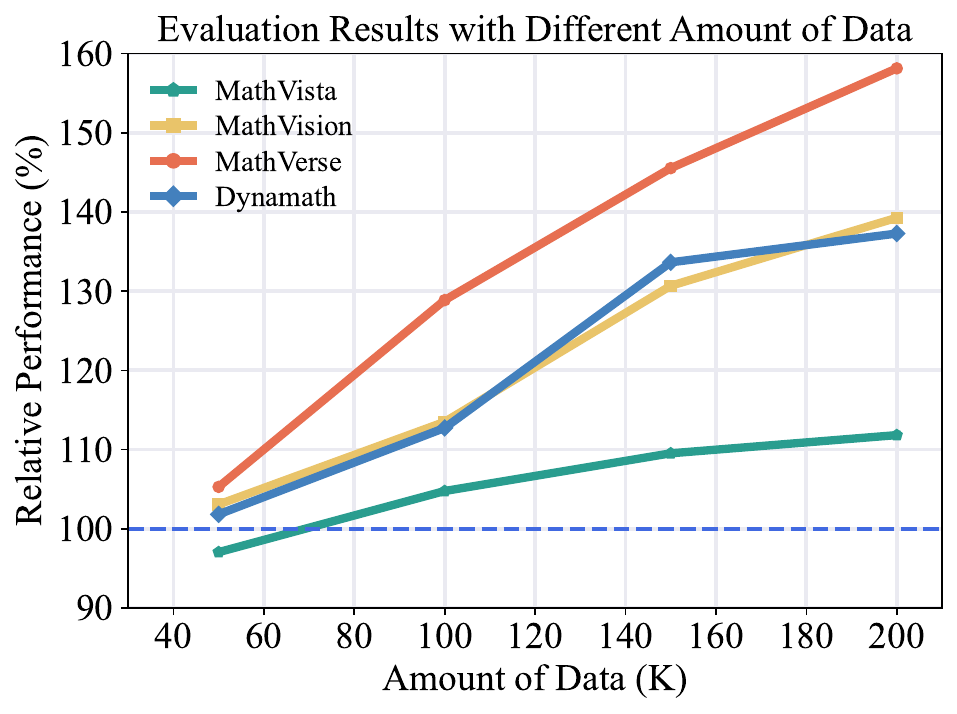}
\caption{\textbf{Ablations on the amount of training data}. \name benefits from data scaling, continually improving the reasoning capabilities.}
\label{fig:data_scaling}
\end{figure}

\subsection{Main Results on Visual Reasoning}
\label{sec:main_result}
The results in Table~\ref{tab:main-exp} demonstrate the effectiveness and generalizability of \name across multiple visual reasoning benchmarks. Notably, our model achieves a 16.7\% improvement over QVQ-72B-Preview on MathVision and a 17.6\% gain on MathVerse, highlighting enhanced reasoning capabilities. Unlike conventional datasets where textual descriptions may include implicit visual cues, MathVerse is an all-around visual math benchmark specifically designed for equitable, in-depth evaluation of MLLMs. The significant gains on MathVerse underscore the significance of \mame, given the benchmark’s unique design principles.
Furthermore, the \mame-7B model, despite its smaller size compared to counterparts, achieves competitive performance, even outperforming leading MLLMs in multiple cases. This demonstrates the robustness of our methodology even with more compact model configurations. Beyond task-specific visual reasoning, we extend our evaluation to general reasoning benchmarks (\eg, MathVista), where \name consistently delivers strong performance, achieving a 3.6\% improvement over the original Qwen2-VL-72B model. These results emphasize \mame’s ability to excel in tasks requiring both perception and reasoning. Collectively, the findings indicate that \name not only advances performance in specialized visual reasoning tasks but also offers substantial benefits in broader application scenarios.

\subsection{Further Analysis}
\label{sec:further_analysis}
In this section, we conduct comprehensive experiments to evaluate the design choices of \mame, emphasizing the key contributions of our approach. We also present a case study to further illustrate the qualitative effectiveness of \mame.

\begin{table}[t]
    \centering
    \caption{\textbf{Ablations on Token Compression.}}
    \resizebox{1.0\linewidth}{!}{
	\begin{tabular}{l|cccc}
        \toprule
		 Method&  MathVista & MathVision &  MathVerse & Avg  \\ 
         \midrule
		 TVC Baseline&  68.3 & 21.5 & 39.6 & 43.1 \\
		 + 2x2 Avg Pooling&  67.8 & 22.9 & 38.3 & 43.0  \\
		 + 4x4 Avg Pooling&  68.1 & 22.7 & 38.9 & 43.2  \\
         \bottomrule
	\end{tabular}
    }
    \label{tab:ablation_token_compression}
\end{table}

\noindent\textbf{Effectiveness of TVC system.} 
To evaluate the effectiveness of the \name system, we conduct comprehensive ablation experiments on various components using the Qwen2-VL-7B as the Baseline. We begin by performing supervised fine-tuning on the Qwen2-VL-7B model with the data from Section~\ref{sec:data_pipeline}, referred to as Vanilla - Direct SFT. Furthermore, we apply the DVR training approach outlined in Section~\ref{sec:TVC}, which increases the focus on the visual information in the training data, enabling the model to implicitly learn visual conditioning capabilities. Additionally, during the testing phase, we experiment with resetting the visual KV cache midway through the reasoning process, and after compressing the visual tokens, we add them to the end of the reasoning steps. This strategy allows the model to further observe the image content during its thought process, mitigating the visual forgetting and suppressing hallucinations.

\begin{figure*}[t]
\centering
\includegraphics[width=1.0\linewidth]{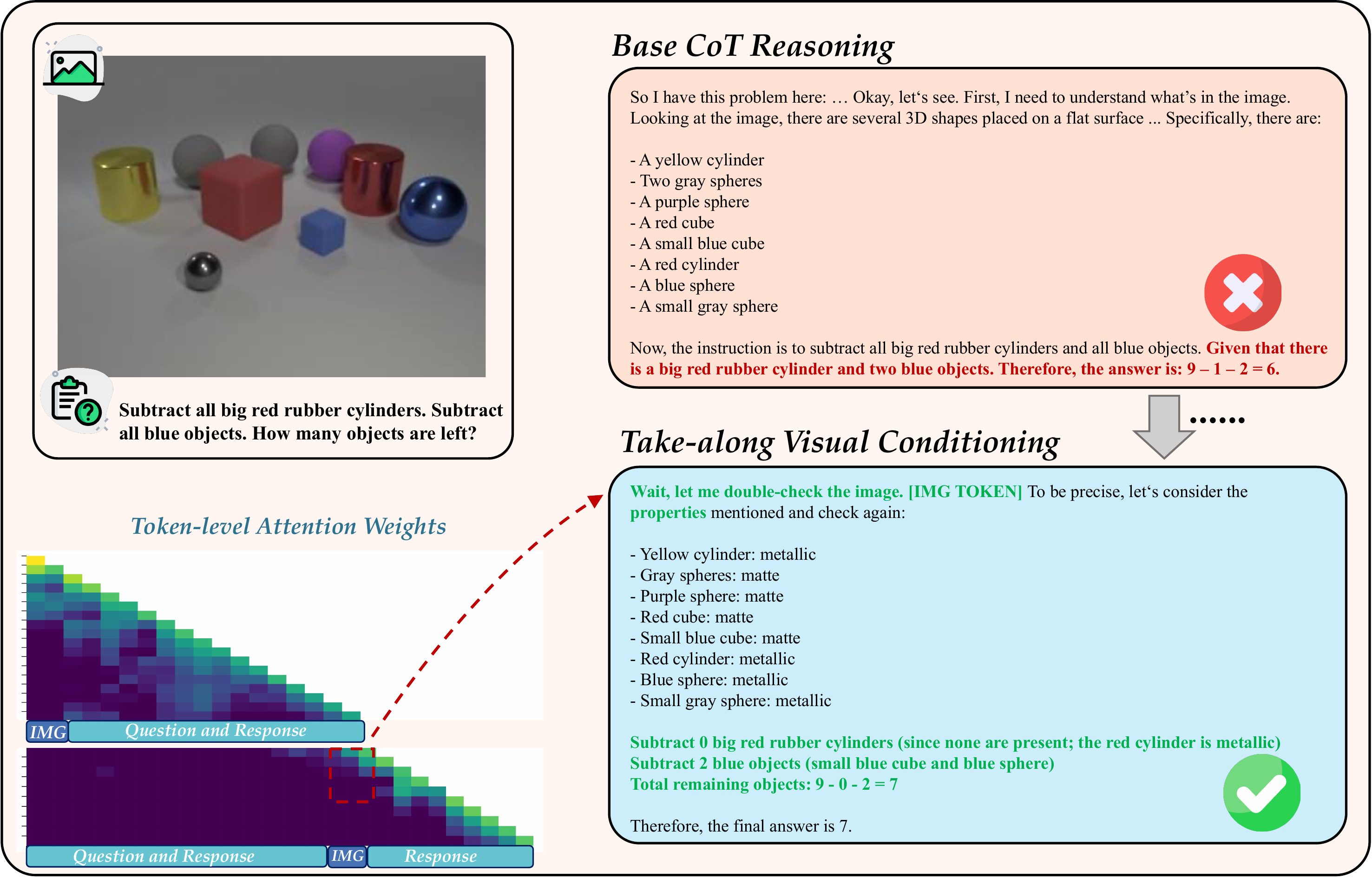}
\caption{\textbf{Case Study of TVC}. \name effectively re-examines the image during the reflection process to correct mistakes, guiding the model to the correct answer.}
\label{fig:case_study}
\end{figure*}

As demonstrated in Table~\ref{tab:ablation_component}, the results highlight that the \name system is crucial for enhancing visual reasoning capabilities. When using only the SFT training data without the DVR strategy in TVC (Vanilla - Direct SFT), improvements in visual reasoning tasks are limited, as the model lacks the ability to reflect on visual information. Furthermore, models trained with the DVR strategy alone still produce sub-optimal results, underscoring the importance of a comprehensive approach that integrates PVC and DVR. The contributions of PVC and DVR are relatively balanced in enhancing the reasoning capabilities.
To further validate the effectiveness of our visual token compression, we conduct experiments with different pooling methods. As shown in Table~\ref{tab:ablation_token_compression}, the TVC Baseline represents the method without image compression. We observe that the use of pooling methods has little impact on the model's capabilities. Utilizing 4x4 average pooling for compression not only enhances the model's inference efficiency but also achieves a slight performance improvement.

\noindent\textbf{Data Scaling Law.} To validate the scalability of our approach, we also conduct a data scaling law experiment for training. As shown in Figure~\ref{fig:data_scaling}, we compare models across various data size: 50K, 100K, 150K, and 200K samples, and present the performance relative to the base model (Qwen2-VL-7B). Our findings show that as the dataset size increases, the model's performance continues to improve. However, it is evident that with a small dataset, the model's reasoning ability cannot reach an optimal level solely through SFT. Therefore, data scaling proves highly effective during SFT training, and the results indicate that \name benefits from increased data.

To prevent underfitting from becoming a performance bottleneck, we increase the number of training epochs with smaller datasets, which further improves model performance. This indicates that insufficient data or epochs lead to undertrained models, making it difficult to learn long-chain reasoning. Increasing both the data and training epochs can effectively alleviate this issue.

\subsection{Case Study}
\label{sec:case_study}
We present a comprehensive case study in Figure~\ref{fig:case_study} to illustrate the improvements of our TVC approach. Specifically, we provide an example that highlights advancements in the reasoning process. In this example, the model is tasked with carefully observing the objects within an image and, after eliminating certain shapes and attributes, providing the count of the remaining objects. During base CoT reasoning, the model fails to check the attributes of each object and only focuses on their shapes, leading to an incorrect final answer. Subsequently, the model learns to re-focus on the image and carefully describe the attributes of each object in detail. This process allows the model to identify the issue in its previous reasoning and provide the correct answer.  \looseness=-1

\section{Conclusion}
In this paper, we introduce Take-along Visual Conditioning (TVC), a novel strategy designed to enhance the reasoning capabilities of MLLMs by addressing the challenge of visual attention degradation during long-chain reasoning. By dynamically shifting the image input to critical stages of reasoning and compressing redundant visual tokens, we ensure that the model maintains focus on the visual information throughout the process. Our extensive evaluation on several mathematical reasoning benchmarks demonstrates the effectiveness of TVC in improving multimodal reasoning, providing a robust approach to equip MLLMs with better visual grounding for complex tasks.

\section*{Acknowledgments}
This work is partially supported by National Key R\&D Program of China (2024YFE0202800), NSFC (62376118),  Key Program of Jiangsu Science Foundation (BK20243012), CCF-Tencent Rhino-Bird Open Research Fund (RAGR20240101), AMED (JP25wm0625405), and Collaborative Innovation Center of Novel Software Technology and Industrialization.

\section*{Limitations}
\label{sec:limitations}
Despite advancements, our method may still exhibit several limitations.
First, for highly complex reasoning tasks requiring sophisticated analytical capabilities, simply increasing visual revisits proves insufficient. In contrast, it is crucial to enhance the model's inherent reasoning capacity. Second, our method assumes the availability of delayed visual processing, making it potentially unsuitable for real-time applications requiring instantaneous visual feedback, such as robotic navigation or time-sensitive decision-making scenarios.

\bibliography{acl}

\newpage
\clearpage
\appendix

\section{Related Work}
\label{sec:related}

\noindent\textbf{Multimodal Large Language Models.} 
Multimodal Large Language Models (MLLMs)~\citep{li2023llava,liu2024llavanext,sun2025parrot,Qwen2VL,lu2024ovis,mckinzie2024mm1,sun2025mos,sun2025pilot,fu2025speculative,dong2024insight} integrate vision encoders~\cite{clip,siglip} with LLMs~\cite{llama-3,qwen-lm}, endowing them with robust capabilities across a wide range of domains. These include general visual understanding\citep{gpt-4o,li2024llava}, mathematical reasoning~\citep{shi2024math,gao2023g}, and answering college-level questions~\citep{Internvl}, demonstrating their versatility in real-world tasks. The rapid advancements in open-source models have also spurred the development of proprietary models, such as GPT-4o~\citep{gpt-4o}, Gemini~\citep{team2023gemini, Gemini1.5}, Qwen2-VL-MAX~\citep{Qwen2VL}, and Claude3~\citep{Claude}. These models have demonstrated remarkable performance in both evaluation benchmarks and practical applications, solidifying their position at the forefront of AI research and deployment.

\noindent\textbf{Reasoning with MLLMs.} 
Recent advancements in MLLMs have significantly enhanced performance in reasoning tasks across both text and multimodal scenarios~\citep{openai2024o1, deepseek2024r1, alibaba2024qvq,peng2025lmm}. Current methods primarily rely on CoT~\citep{wei2022chain} to train MLLMs for step-by-step reasoning. Data-driven approaches include Math-LLaVA~\citep{shi2024math}, which introduced the MathV360K dataset, and MAmmoTH-VL~\citep{guo2024mammothvl}, which curates a large-scale multimodal CoT dataset in a scalable manner.
Another line of research explores vision-text alignment. MAVIS~\citep{zhang2024mavis} fine-tunes a math-specific vision encoder with curated caption data, while Math-PUMA~\citep{zhuang2024math} leverages the Kullback-Leibler (KL) divergence of next-token prediction distributions for modality alignment.
In a different paradigm, MLLMs act as coordinators, utilizing external tools such as LLMs, web search engines, and computer programs for complex reasoning. Chameleon~\citep{lu2023chameleon} orchestrates tool-call sequences, and Visual Sketchpad~\citep{hu2024visual} enables models to generate visual sketches to aid reasoning.

\section{More Details of Reasoning Dataset}
\label{sec:prune_dataset}
In this section, we provide a detailed description of dynamic token truncation and reflection word pruning in the process of constructing the reasoning dataset. We also provide detailed information about the training data in Table~\ref{tab:train-data}.
\subsection{Dynamic Token Truncation}
To further improve the dataset quality, we analyze the distribution of token lengths after the answer-centric rejection sampling. We find that many samples are close to the maximum token limit, and manual checks show that these long reasoning chains often have problems—such as logical errors, mistakes in multi-step calculations, and reliance on shortcuts that don’t work in general cases (\eg, substituting specific values). Motivated by the correlation between extreme token lengths and compromised solution quality, we implement adaptive truncation thresholds to keep the answers within the 200-8000 token range. This dynamic filtering not only eliminates the invalid cases (overly verbose or terse responses) but also enhances the overall quality of the data. The final length distribution matches how human experts solve problems and keeps the important reasoning steps intact.

\subsection{Reflection Word Pruning}
Our analysis reveals a critical flaw in distilled reasoning chains: excessive metacognitive loops caused by uncontrolled reflection markers (\eg, `Alternatively,' `Wait'), which led to performance degradation through infinite loops or ungrounded speculation. Term frequency analysis of reflection density shows a heavy-tailed distribution—95\% of samples contained fewer than 10 reflection markers per chain, while 1\% exhibited over 50 markers, with this extreme group strongly correlating to hallucination rates. To address this, we introduce a reflection token quota system that automatically prunes samples exceeding 25 reflection markers while retaining core reasoning logic using semantic-aware span detection. As shown in Figure~\ref{fig:data_pipeline}, this approach significantly reduced infinite-loop instances in validation tasks while improving answer accuracy. The refined reflection pattern mirrors expert human problem-solving strategies, wherein targeted self-correction enhances, rather than disrupts, the continuity of the reasoning process. \looseness=-1

\begin{figure*}[t]
\centering
\includegraphics[width=1.0\linewidth]{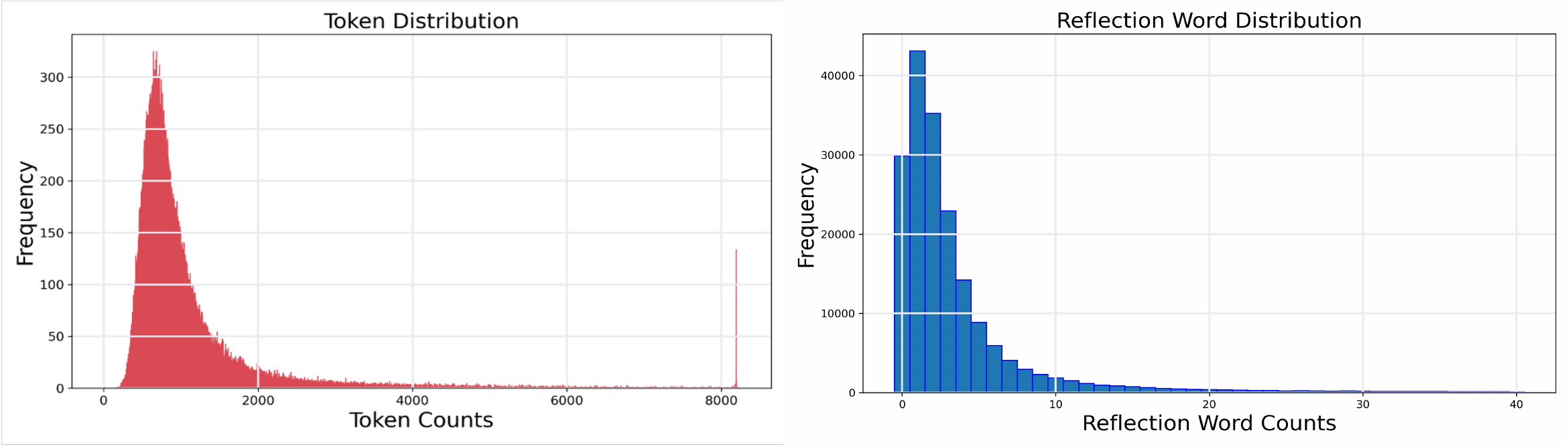}
\caption{The token and reflection word distribution of the long-chain reasoning dataset.}
\label{fig:distribution}
\end{figure*}

\begin{table}[t]
  \small
  \caption{The detailed training hyperparameters.}
  \label{tab:train_detail}
  \centering
  \begin{tabular}{l|cc}
    \toprule
    Config & SFT \\
    \midrule
    Deepspeed & Zero3 \\
    Epoch & 5 \\
    Warmup Ratio & 0.1 \\
    Max Grad Norm & 1.0 \\
    Optimizer & AdamW \\
    Learning rate & 2e-5 \\
    Learning rate scheduler & Cosine \\
    Text max length & 8192 \\
    Batch size per GPU &  1 \\
    Gradient Accumulation Steps &  4 \\
    GPU & 64×H20-96G  \\
    Precision & Bf16 \\
    \bottomrule
  \end{tabular}
\end{table}

\section{Discussion}
\label{sec:discussion}
In this work, we investigate the phenomenon of visual information forgetting in MLLMs during long-chain reasoning. Through comprehensive analysis and experiments, we show that as reasoning chains progressively lengthen, models exhibit a gradual deterioration in retaining visual inputs, ultimately undermining their multimodal reasoning capabilities and exacerbating hallucination issues. To address this challenge, we propose a take-along visual conditioning mechanism that enables models to dynamically revisit visual inputs during reasoning steps, thereby enhancing content fidelity throughout the inference process. 

\begin{figure*}[t]
    \centering
    \includegraphics[width=1.0\linewidth]{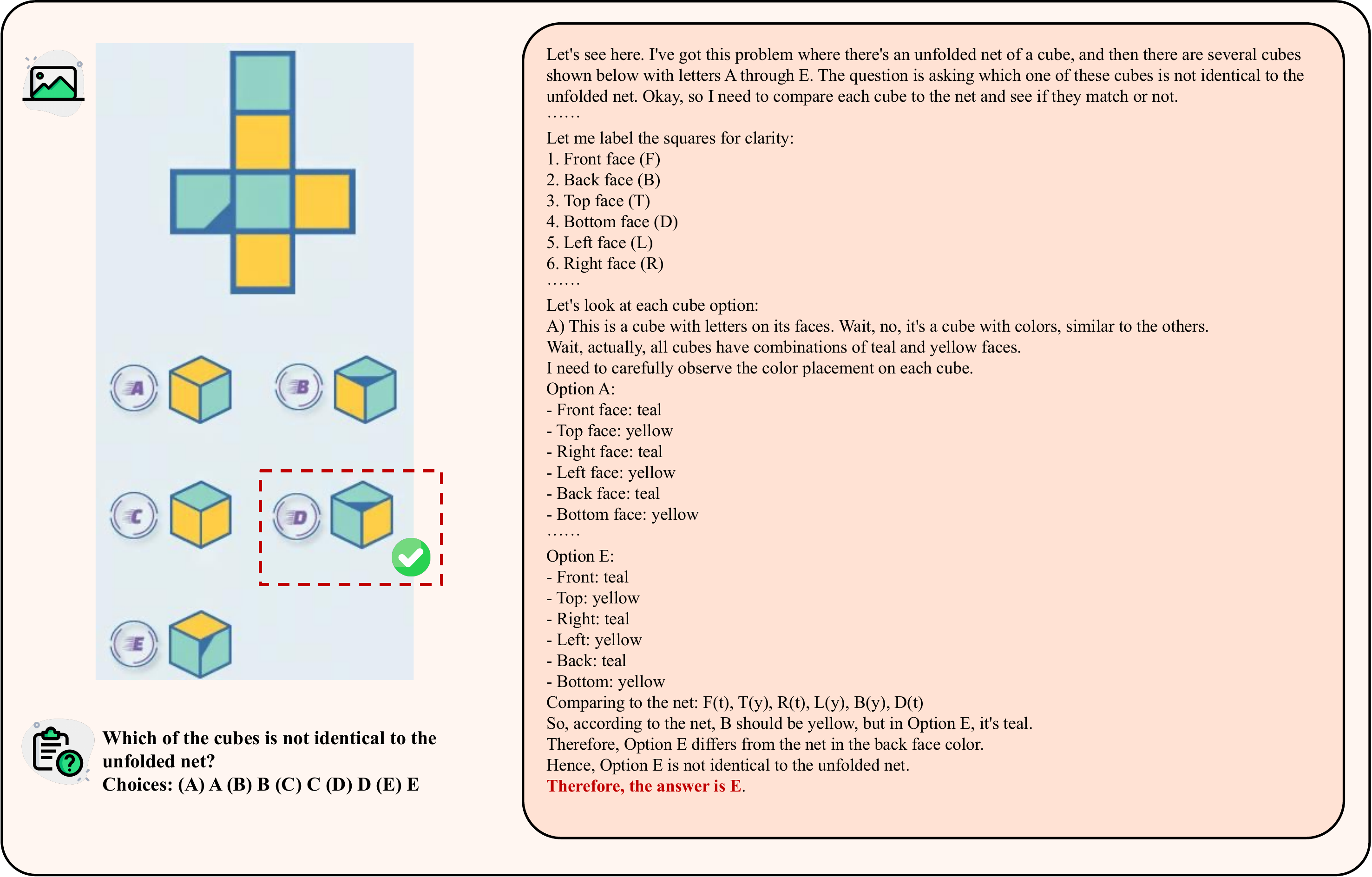}
    \caption{Qualitative Results of \mame.}
    \label{fig:limitation}
\end{figure*}

However, as illustrated in Figure~\ref{fig:limitation}, we acknowledge several limitations. First, for highly complex reasoning tasks requiring sophisticated analytical capabilities, simply increasing visual revisits proves insufficient. In contrast, it is crucial to enhance the model's inherent reasoning capacity. Second, our method assumes the availability of delayed visual processing, making it potentially unsuitable for real-time applications requiring instantaneous visual feedback, such as robotic navigation or time-sensitive decision-making scenarios.

Our work represents an initial exploration into mitigating visual forgetting in extended multimodal reasoning chains. We envision future research directions including: (1) Developing hybrid architectures that synergistically enhance both visual retention and intrinsic reasoning capabilities; (2) Investigating adaptive attention mechanisms for real-time multimodal applications; (3) Exploring curriculum learning strategies to progressively strengthen long-chain reasoning capacities. We hope this foundational study will inspire further advances in understanding and improving multimodal reasoning systems for complex real-world applications.

\begin{table}[t]
\caption{Details on the \mame's training data, which is derived from publicly available datasets.}
\centering
\scalebox{0.98}{
    \begin{tabular}{@{}cccc@{}}
    \toprule
    Datasets & Samples\\ 
    \midrule
    MathV360K~\citep{shi2024math} & 221K\\ 
    Geo170K~\citep{gao2023g} & 22K\\ 
    LLaVA-OneVision~\citep{li2024llava} & 97K\\ 
    Cambrian-1~\citep{tong2024cambrian} & 1K\\ 
    \bottomrule
    \end{tabular}
}
\label{tab:train-data}
\end{table}

\end{document}